\documentclass[lettersize,journal]{IEEEtran}
\usepackage{amsmath,amsfonts}
\usepackage{algorithmic}
\usepackage{algorithm}
\usepackage{array}
\usepackage{hyperref}
\usepackage{booktabs}
\usepackage{textcomp}
\usepackage{stfloats}
\usepackage{url}
\usepackage{verbatim}
\usepackage{graphicx}
\usepackage{color}
\usepackage{cite}
\usepackage{multirow}
\usepackage{tabularray}
\usepackage{subcaption}
\usepackage{tabularx}
\usepackage{threeparttable}
\begin{document}
\title{GrabDAE: An Innovative Framework for Unsupervised Domain Adaptation Utilizing Grab-Mask and Denoise Auto-Encoder}

\author{
        Junzhou Chen, Xuan Wen, Ronghui Zhang, Bingtao Ren, Di Wu, Zhigang Xu, Danwei Wang, $\textit{Fellow, IEEE}$
        \thanks{This project is jointly supported by National Natural Science Foundation of China (Nos. 61003143, 52172350), Guangdong Basic and Applied Research Foundation (Nos.2021B1515120032, 2022B1515120072), Guangzhou Science and Technology Plan Project (No.2024B01W0079), Nansha Key RD Program(No.2022ZD014), Science and Technology Planning Project of Guangdong Province (No.2023B1212060029). \textit{(Corresponding author: Ronghui Zhang.)}}
        \thanks{Junzhou Chen, Xuan Wen, Ronghui Zhang are with the Guangdong Provincial Key Laboratory of Intelligent Transport System, School of Intelligent Systems Engineering, Sun Yat-sen University, Guangzhou 510275, China (e-mail: chenjunzhou@mail.sysu.edu.cn, wenx73@mail2.sysu.edu.cn, zhangrh25@mail.sysu.edu.cn).}
        \thanks{Bingtao Ren is with the School of Transportation Science and Engineering, Beihang University and with State Key Lab of  Intelligent Transportation System, Beijing 100191, China. (e-mail: renbt1706@buaa.edu.cn).}
        \thanks{Di Wu is with the School of Computer Science and Engineering, Sun Yat-sen University, Guangzhou, 510006, China, and also with the Guangdong Key Laboratory of Big Data Analysis and Processing, Guangdong 510006, China (e-mail:wudi27@mail.sysu.edu.cn).}
        \thanks{Zhigang Xu is with the School of Information Engineering, Chang'an University, Xi'an 710064, Shaanxi, China (e-mail: xuzhigang@chd.edu.cn).}
        \thanks{Danwei Wang is with the School of Electrical and Electronic Engineering, Nanyang Technological University, Singapore 639798 (e-mail: edwwang@ntu.edu.sg).}
        \thanks{This paper was submitted to IEEE Transactions on Multimedia (TMM) on May 25, 2024.}
}


\maketitle

\maketitle

\begin{abstract}

Unsupervised Domain Adaptation (UDA) aims to adapt a model trained on a labeled source domain to an unlabeled target domain by addressing the domain shift. Existing Unsupervised Domain Adaptation (UDA) methods often fall short in fully leveraging contextual information from the target domain, leading to suboptimal decision boundary separation during source and target domain alignment. To address this, we introduce GrabDAE, an innovative UDA framework designed to tackle domain shift in visual classification tasks. GrabDAE incorporates two key innovations: the Grab-Mask module, which blurs background information in target domain images, enabling the model to focus on essential, domain-relevant features through contrastive learning; and the Denoising Auto-Encoder (DAE), which enhances feature alignment by reconstructing features and filtering noise, ensuring a more robust adaptation to the target domain. These components empower GrabDAE to effectively handle unlabeled target domain data, significantly improving both classification accuracy and robustness. Extensive experiments on benchmark datasets, including VisDA-2017, Office-Home, and Office31, demonstrate that GrabDAE consistently surpasses state-of-the-art UDA methods, setting new performance benchmarks. By tackling UDA's critical challenges with its novel feature masking and denoising approach, GrabDAE offers both significant theoretical and practical advancements in domain adaptation.

\end{abstract}


\section{Introduction}
\IEEEPARstart{I}{n} recent years, machine learning and deep neural networks have revolutionized visual classification tasks, achieving remarkable accuracy and efficiency. Yet, their effectiveness typically hinges on two pivotal assumptions: the need for extensive labeled data for model training and the prerequisite that the data employed for testing or application must be consistent in distribution with the training dataset\cite{deng2021informative,wang2022information,lu2021discriminative,shermin2020adversarial}. A model's success in accurately classifying data is contingent upon these conditions. Notably, a disparity in distribution between training and testing datasets, a challenge often referred to as domain shift, frequently leads to suboptimal performance during the testing or application phase\cite{hoyer2023domain,shermin2020adversarial}.

\begin{figure}
\centering
\includegraphics[width=3.5in]{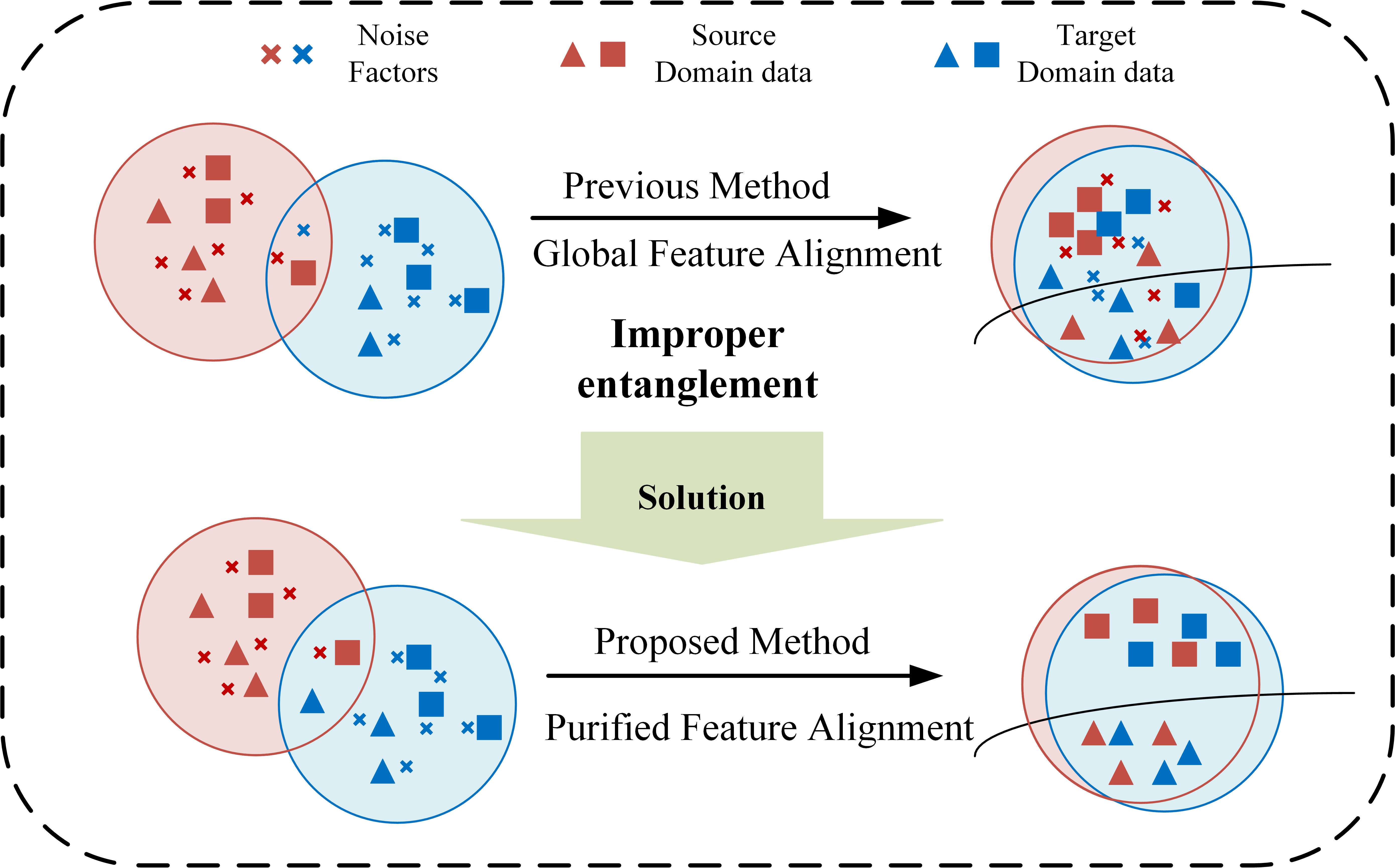}
\caption{Illutration of the issues method and our solution in proposed method. In the task of classification, Traditional Domain Adaptation learning embeds various noise factors, causing these features prone to be wrongly aligned between two domains (improper entenglement).  Based on the pure semantic information from upstream, GrabDAE further boosts the category-level feature alignment by employing DAE module to reconstructing features.}
\label{Denoise}
\end{figure}

Significant manpower and resources are required for the collection and annotation of data, which underpins the training of models for visual tasks. The promise of training models on small datasets or those directly relevant to the target visual task is thus highly valued\cite{sener2016learning}. However, the issue of domain shift poses a significant limitation on model's ability to extend applicability to the target domain. As a highly competitive proposal, unsupervised Domain Adaptation (UDA) can effectively overcome this limitation\cite{hoyer2023mic,sun2017correlation}.

The primary objective of UDA is to transfer knowledge from a labeled source domain to an unlabeled target domain \cite{yang2023tvt,xu2021cdtrans} and leverage these insights for applications in the target domain \cite{zhang2018collaborative,baktashmotlagh2013unsupervised}. Prevailing UDA approaches can be broadly categorized into two main strategies: learning domain-invariant features and aligning data distributions between the two domains\cite{kang2020contrastive,saito2020universal,9108582}.

Despite advancements in deep learning and machine learning, challenges related to data dimensionality and complexity remain central. The growth in dataset size and complexity often outpaces the ability of conventional methods to capture inherent data structures, potentially stifling model performance\cite{rangwani2022closer}. Additionally, the data features frequently suffer from the detrimental effects of noise during model propagation, impacting both model performance and reliability\cite{pinheiro2018unsupervised,long2014transfer,rangwani2022closer}. Enhancing data representation capabilities, reducing data dimensionality, and leveraging data-based insights become imperative in UDA tasks.

Moreover, GrabDAE incorporates a Denoise Auto-Encoder (DAE) module that plays a critical role in feature purification. By reconstructing features from a higher-dimensional space, this module effectively filters out noise and irrelevant features, ensuring that the model's learning process is driven by the most informative aspects of the data. This mechanism of feature purification is instrumental in achieving superior classification accuracy in the target domain. Besides, the reconstruction process in DAE can be viewed as an implicit regularization mechanism. By adding noise to the input features and forcing the model to denoise and reconstruct them, DAE encourages the model to focus on learning the underlying, meaningful structure of the data rather than overfitting to noise or domain-specific artifacts. This process effectively reduces the model’s reliance on spurious features from the source domain and promotes the learning of more generalized and robust representations that are applicable across domains. As illustrated in Figure. 1, traditional domain adaptation learning processes are shown to embed various noise factors, leading to misalignment between domains. In contrast, GrabDAE leverages the DAE module to reconstruct features based on pure semantic information from upstream, enhancing the alignment of category-level features and showcasing the model's advantages.

The primary contributions of this study are multifaceted, encompassing four advancements:

\begin{itemize}
    \item[\textbf{(i)}] \textbf{Comprehensive Integration of Adaptive Techniques:} \textbf{GrabDAE} represents a breakthrough in domain adaptation frameworks by its comprehensive integration of adversarial training, metric learning, and self-supervised learning. This unified framework directly addresses the challenge of misaligned feature distributions across domains, significantly enhancing the extraction of domain-invariant features and improving model performance across diverse datasets.
    
    \item[\textbf{(ii)}] \textbf{Strategic Enhancement with the Grab-Mask Module:} The introduction of the \textbf{Grab-Mask module} in GrabDAE provides a novel mechanism to mitigate the impact of environmental noise and variability in the target domain. By focusing on core, domain-relevant features, it strengthens the model’s resilience to domain shifts, leading to improved predictive accuracy in unsupervised domain adaptation.
    
    \item[\textbf{(iii)}] \textbf{Innovative Use of Denoise Auto-Encoder for Feature Stability:} The \textbf{Denoising Auto-Encoder (DAE)} is designed to enhance feature stability by simulating the process of information transmission, addressing noise and variability in feature extraction. This innovation refines feature compression and decompression, improving the stability and efficiency of domain adaptation.

    \item[\textbf{(iv)}] \textbf{Benchmark-Setting Performance:} \textbf{GrabDAE} demonstrates consistent improvements over existing UDA methods in extensive evaluations on benchmark datasets, including VisDA-2017, Office-Home, and Office31. The results establish new performance baselines, highlighting GrabDAE’s robustness and practical value for a wide range of visual classification tasks.
\end{itemize}

\begin{figure*}
\centering
\includegraphics[width=5.5in]{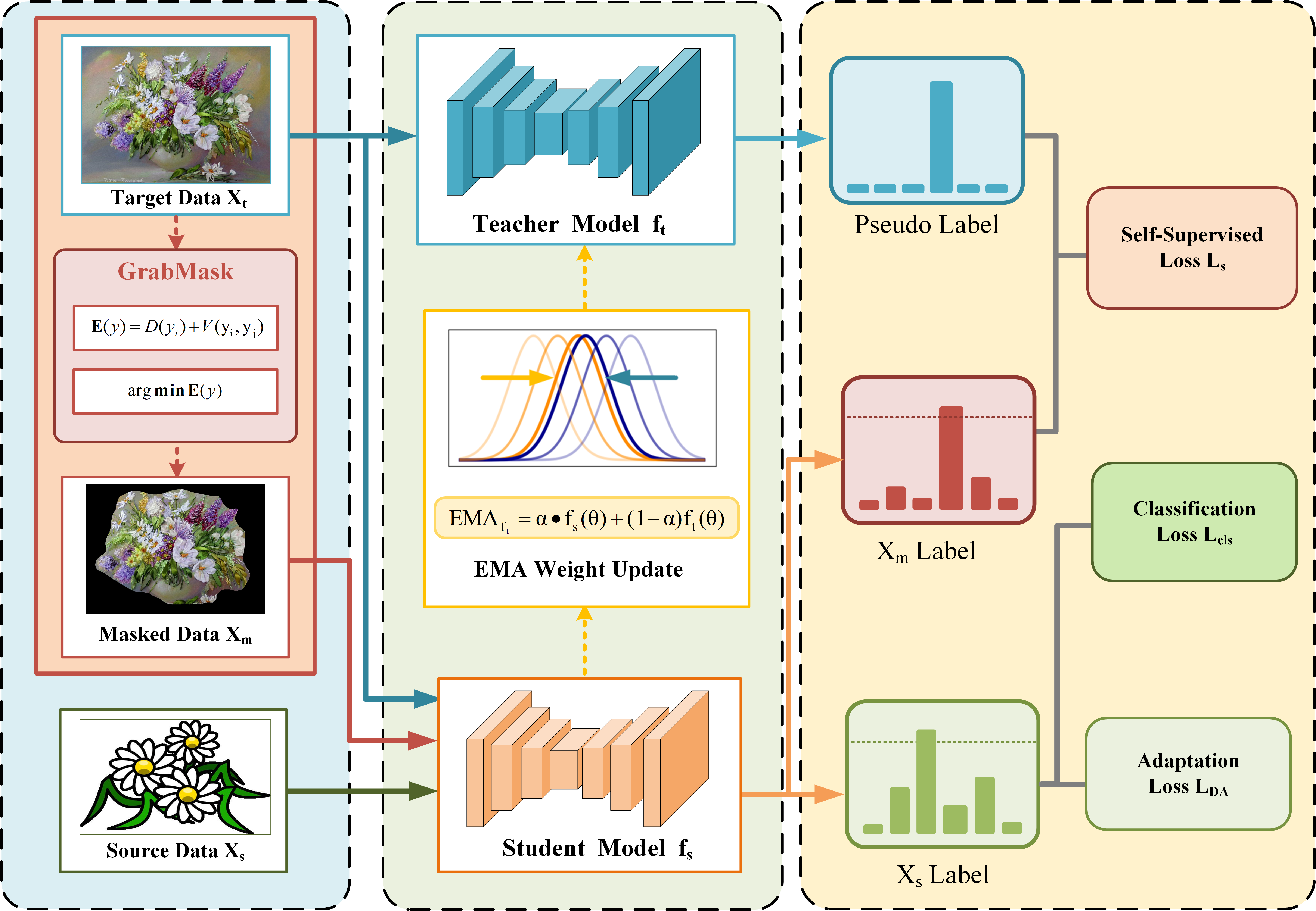}
\caption{Overview of the GrabDAE.In UDA, a network is typically trained with a supervised loss on the source domain data\cite{venkateswara2017deep} (green) and an unsupervised adaptation loss on the target domain data\cite{venkateswara2017deep} (blue). GrabDAE enforces the consistency between predictions of masked target images(red) and pseudo labels(blue) that are generated based on the complete image by teacher model.}
\label{overview}
\end{figure*}

\section{Related work}

\textit{Masking Techniques}. The technique of masking input data before prediction was initially employed in self-supervised learning tasks within natural language processing. Recently, this concept has been effectively integrated into self-supervised pre-training methodologies within the realm of computer vision. By providing partially masked images to the network and training it to reconstruct the attributes of the masked regions, this approach has shown substantial promise in enhancing model understanding of image content\cite{germain2015made,hoyer2023mic,peng2024unsupervised}. The MAE proposed by He et al.\cite{he2022masked} underscores the redundancy of pixels in images and the potential of masking techniques to improve model efficiency and focus. Moreover, masking images before model training has become a prevalent practice, aiding in better capturing contextual cues within images for more accurate predictions\cite{pinheiro2018unsupervised,hoyer2023domain}.

In the task of classification in UDA, masking techniques, such as the MIC method, involve masking random blocks of target images to enforce network learning of context for semantic segmentation of the entire image. This method enables models to better utilize contextual cues by focusing on learning the semantics of the masked regions\cite{hoyer2023mic}. Our method, GrabDAE, diverges from previous works by employing a directed masking approach. Rather than treating each image region equally, it prioritizes regions critical for the visual task of classification, using traditional data segmentation methods like GrabCut\cite{rother2004grabcut} to refine target domain data. This approach ensures that the model focuses on areas of interest within the target domain, thereby eliminating irrelevant factors and facilitating more effective learning of semantic clues.

\textit{Unsupervised Domain Adaptation (UDA)}. UDA endeavors to train models using labeled data with the aim of generalizing to the unlabeled target data\cite{yang2023tvt,zhu2023patch}. 
In general, UDA is conventionally divided into two sequential steps. The first step involves learning features from the source data with label, and the second step involves transforming the features of source data into data features usable in the target domain\cite{ghifary2016deep,lu2021discriminative}. This adaptation facilitates the deployment of the classifiers trained on the source domain in the target domain. Current research in domain adaptation primarily focuses on methods to minimize domain variance and align distict domain features into a shared space\cite{pan2009survey,odena2017conditional,long2017deep,sun2016deep} or employ adversarial learning to aid the model in better adapting to diverse data distributions\cite{jing2022adversarial,lu2021discriminative,shermin2020adversarial,li2019cycle,ma2019deep}.

Adversarial training has emerged as a promising technique for addressing the challenges presented by domain variation\cite{jing2022adversarial,lu2021discriminative,shermin2020adversarial,li2019cycle,ma2019deep}. In this context, the problem is conceptualized as a minimax game involving feature learning and a domain classifier\cite{zhu2023patch,zheng2024semantics}. Feature learning aims to confuse and learn from cross-domain features, while the discriminator's task is to differentiate whether data features are from the target or source domain\cite{zhu2020deep,kang2020contrastive}. Generally, after passing through the feature extraction module, the data's features are separately fed into a classifier and a discriminator for domain adaptation.

Despite these advances, most domain adaptation methods cannot adequately address the relationship between the data distribution inherent to the target domain and its corresponding decision boundaries, potentially leading to diminished discriminability within the target domain\cite{dosovitskiy2020image,kang2020contrastive,ding2022source,zheng2024semantics,yue2023make,li2023adjustment}. Our approach seeks to overcome this limitation by incorporating self-supervised learning through reconstruction loss and minimizing consistent loss between the pseudo labels and the predictions of masked target domain data. This strategy significantly improves the recognition of inter-class disparities, consequently improving the accuracy of our model.

\textit{Denoise Auto-Encoder (DAE)}. The Denoising Auto-Encoder (DAE) consists of an encoder that maps noisy inputs to a latent representation space and a decoder that reconstructs the corrupted inputs from this representation \cite{yadav2022integrated}. Much like how humans can recognize partially occluded or damaged images, a well-trained DAE mimics this ability by ensuring that the hidden layer representation remains consistent for both intact and damaged inputs. This consistency allows the DAE to effectively reconstruct clean input signals from corrupted or occluded inputs, showcasing the model's robustness and versatility in handling diverse input conditions \cite{venkateswara2017deep}.
DAEs are widely used in self-supervised pretraining within computer vision, focusing on reconstructing inputs from latent representations and highlighting the most informative features for classification \cite{kan2015bi,ghifary2016deep}. Our novel adaptation of the DAE leverages masked images to facilitate the learning of domain-adaptive contextual relationships, thereby significantly improving the model's ability to interpret structural information within the target domain\cite{ding2022source,peng2024unsupervised}. By introducing Gaussian noise perturbations at the encoder stage, our approach ensures that feature vectors across different domains conform to a Gaussian distribution, promoting effective transfer learning and enhancing domain alignment.

\begin{figure*}[!t]
\centering
\includegraphics[width=6.8in]{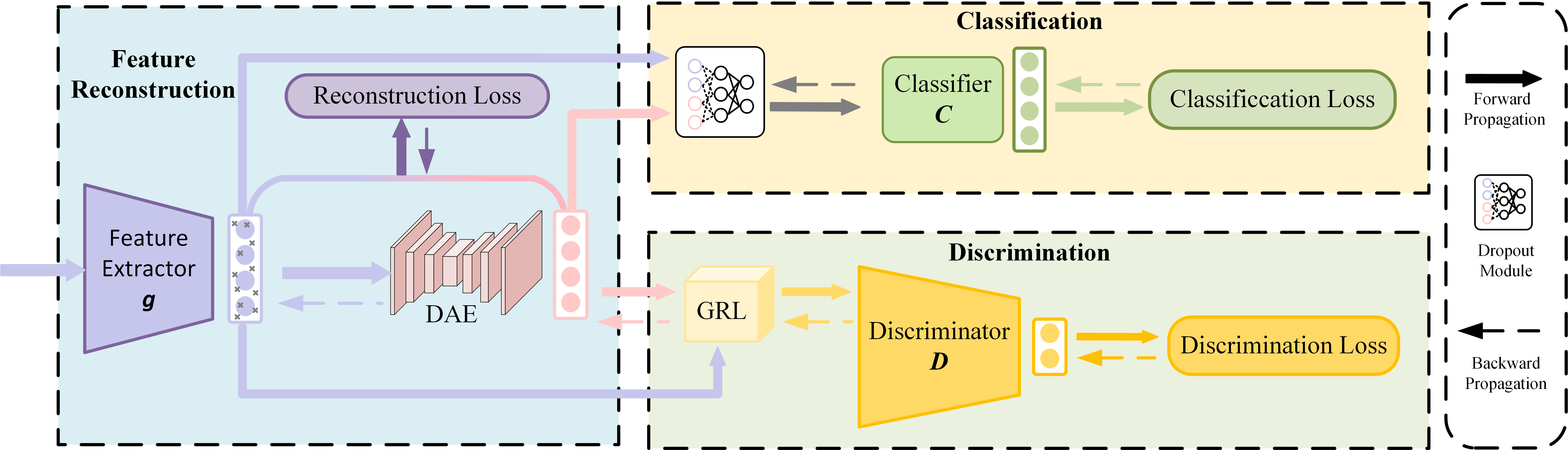}
\caption{Architecture of the model.The model consists of three main components: feature reconstruction (blue), classification (yellow), and discrimination (green). In the feature reconstruction part, data is processed through a feature extractor and a Denoise Auto-Encoder (DAE) to compute a reconstruction loss. In the classification part, both original features and reconstructed features are fed into a classifier via a Dropout layer for classification, computing a classification loss. In the discrimination part, both types of features are input to a discriminator for discrimination, calculating a discriminator loss.}
\label{architecture}
\end{figure*}

\section{Methods}
In this section, we introduce the technical notation of our method for unsupervised domain adaptive image classification and specify the objectives in each considered setting.

\subsection{Unsupervised Domain Adaption}

Our method aims to train an image classification model $f_{s}$ capable of predicting correct labels for $T=\{x_{i}^{t}\}$, achieving similar performance to that on the labeled source domain $S = \{(x_{i}^{s}, y_{i}^{s})\}$.

First, we assume that the two domains possess a shared label scope. In UDA, the training process incorporates unsupervised learning on unlabeled target data $T$ and supervised learning on labeled source data $S$. Hence, the supervised categorical cross-entropy loss is computed solely based on predictions for the labeled data, denoted as $\hat{y}^{S} = f_{s}(x^{S})$.
\begin{equation}
    \mathcal{L}_{cls} = \frac{1}{n_{s}} \sum_{x_{i}\in S } \mathcal{L}_{ce} (\hat{y}^{S}, y^{S})
\end{equation}

Here, labeled source data contributes to learning domain-invariant features, while unlabeled target data facilitates feature alignment across the two domains by leveraging underlying data distribution. The objective is to minimize the distribution discrepancy between source and target domains, improving the model’s performance on the target domain by learning from the source. We achieve this through optimizing the following total loss function:

\begin{equation}
    \min \mathcal{L}_{cls} + \mathcal{L}_{s} + \mathcal{L}_{DA}(x^{s}, x^{t})
\end{equation}

where $\mathcal{L}_{cls}$ denotes the supervised loss for source domain predictions, $\mathcal{L}_{s}$ the self-supervised loss for the target domain, and $\mathcal{L}_{DA}$ the domain adaptation loss facilitating feature alignment across domains.

In our framework, Model $f$ distills knowledge from the source domain to minimize the supervised loss $\mathcal{L}_{cls}$. The domain adaptation loss is decomposed as
\begin{equation}
    \mathcal{L}_{DA} = \mathcal{L}_{re} + \mathcal{L}_{D}
\end{equation}

The model first uses the original, unprocessed target domain images to generate pseudo-labels, providing a supervisory signal for the unlabeled data. Following this, the model applies GrabMask to the target images and learns the consistency between these processed images and the pseudo-labels. This process reduces the influence of background noise, allowing the model to better focus on extracting domain-invariant features and mitigating domain differences.

Our GrabDAE framework also employs a Teacher-Student architecture, where pseudo-labels are generated by an Exponential Moving Average (EMA) teacher model. The teacher model makes soft-label predictions based on the original target domain images, ensuring stability and robustness in the pseudo-labels. In the student model, labeled source domain data and two types of target domain data (original and GrabMask-processed) are fed into the model. By computing the classification loss on the source domain and domain adaptation loss, the model's transferability is further optimized. This dual-optimization mechanism enhances the model's performance on the source domain classification task and strengthens its adaptability to target domain data through contrastive learning and consistency regularization.

\subsection{GrabMask}

The Grab-Mask module is a vital component of the proposed GrabDAE framework for unsupervised domain adaptation. The purpose of Grab-Mask is to guide the model to focus on regions of interest and produce accurate classification results, which do not necessarily require distant contextual information.

We leverage Gaussian Mixture Models (GMMs) to perform soft segmentation\cite{riesaputri2020classification}. 
The method involves iteratively refining an initial user-defined foreground and background segmentation by modeling image pixels as a mixture of Gaussian distributions. Initially, we introduce seed points to provide a rough estimate of the foreground and background regions.

Foreground and background distinction is further refined using the Gibbs energy equation:
\begin{equation}
E(y) = D_i(y_i) + V_{i, j}(y_i, y_j)
\end{equation}

where the label $y_i$ is assigned to pixel $i$, $D$ quantifies the cost of assigning label $y_i$ to pixel $i$ based on the intensity value of the pixel, and $V$ measures the cost of assigning different labels to neighboring pixels $i$ and $j$, encouraging neighboring pixels to have the same label. Specifically, the smoothness term $V$ is defined as:
\begin{equation}
    V(i, j) =  \gamma \cdot \exp\left( -\frac{\|z_i - z_j\|^2}{2 \sigma^2} \right) \cdot \mathbb{I}[l_i \neq l_j]
\end{equation}
where:
\begin{itemize}
    \item \( z_i \) and \( z_j \) represent the color vectors of pixels \( i \) and \( j \).
    \item \( \sigma^2 \) is the variance of the color differences in the local neighborhood.
    \item \( \mathbb{I}[l_i \neq l_j] \) is an indicator function that equals 1 if the labels \( l_i \) and \( l_j \) of pixels \( i \) and \( j \) are different, and 0 otherwise.
    \item \( \gamma \) is a constant controlling the strength of the smoothness term.
\end{itemize}

To ensure that the alignment of minimum of the energy function $E$ with a segmentation of superior quality, information derived from the frequency distributions of observed grayscale values is integrated into its formulation.
The iterative minimization process ensures convergence and enhances classification capabilities by directing attention towards objects of interest.

In synthesis, the Grab-Mask module, serves as a gateway for preprocessed, saliency-enhanced image representations through its sophisticated utilization of GMM-based soft segmentation and energy minimization. These representations are subsequently fed into the Unsupervised Domain Adaptation (UDA) process, guiding the model's learning towards the most visually pertinent cues. This not only enhances adaptability to domain shifts but also strengthens overall classification performance.


\subsection{Self-Supervised Learning for UDA}

To improve the adaptation of the model $f_s$ to the specific data, we implement a self-supervised learning approach. This method utilizes pseudo-labels derived from target data, iteratively refining model predictions to better align with these labels. This strategy progressively boosts the model's generalization capabilities towards the target domain $T$, exploiting the consistency between pseudo-labels and the prediction of masked target data to achieve robust adaptation. This iterative process includes generating initial pseudo-labels for target data $T$, using them for self-supervised training\cite{bucci2021self,xiao2021self,sener2016learning,long2015learning}, and applying consistency loss to maintain prediction stability. Over iterations, pseudo-labels are refined, enhancing model adaptation to the target domain. This approach leverages self-supervised learning to minimize dependency on labeled data, making the adaptation process scalable and efficient.

GrabDAE withholds highly correlated information within the samples using GrabMask. We obtain the masked target image $x^{M}$ through GrabMask:
\begin{equation}
    x^{M} = GrabMask(x^{T})
\end{equation}

Subsequently, the prediction for the masked target domain data $\hat{y}^M$ is derived using the student network $f_s$:
\begin{equation}
    \hat y^{M}=f_{s}(x^{M})
\end{equation}

In the target domain, samples lack labels. Self-supervised learning can obtain the advanced features of target domain data by leveraging the labels of $x^{M}$ and the pseudo-labels of $x^{T}$. The self-supervised loss is introduced as:
\begin{equation}
    \mathcal{L}_{s} = \mathcal{L}_{ce}(\hat{y}^{M},p^{T})
\end{equation}

Significantly, the pseudo-label is the prediction of the complete target domain image $x_{T}$ made by a teacher network $f_{t}$.
\begin{equation}
    p^{T} = \mathop{\arg\max}\limits f_{t}(x^{T})
\end{equation}

Throughout the self-supervised training process, the student model $f_{s}$ is trained on masked target data and the teacher model $f_{t}$ dynamically generates pseudo-labels from original Target data. Specifically, the teacher model $f_{t}$ updates itself after each step, incorporating an exponentially weighted average of the student model's weights. Note that no gradients are propagated back into teacher model $f_{t}$.

\begin{figure}
\centering
\includegraphics[width=3.5in]{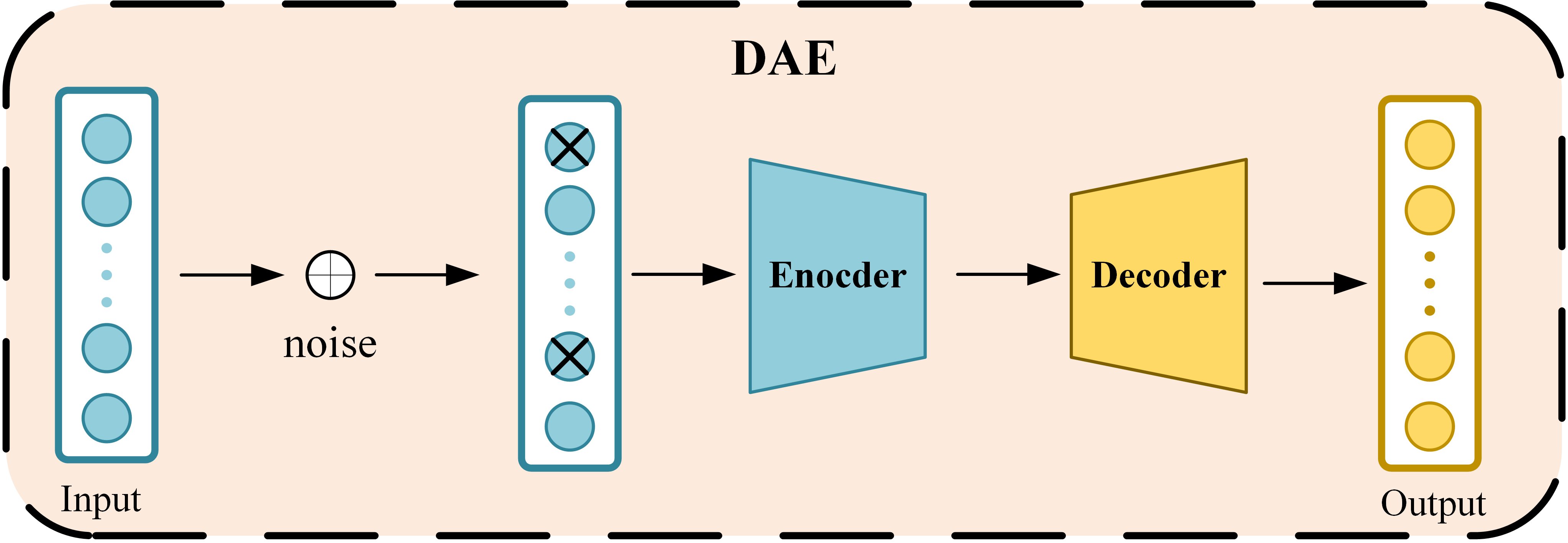}
\caption{Illustration of the Denoise Auto-Encoder (DAE) process within the GrabDAE framework. The figure depicts the sequence of adding Gaussian noise to the input, encoding to a latent representation, decoding to reconstruct the original input, and the final output. This process underscores the DAE's role in feature purification and noise reduction, enhancing the model's adaptability and performance in domain adaptation tasks.}
\label{DAE}
\end{figure}

\subsection{Denoise Auto-Encoder}


In Fig. 3, model is structured in three phases: feature reconstruction, classification, and discrimination. In the feature reconstruction phase, image data is passed through $g$ and the DAE to remove noise and enhance the stability of learned representations. In the classification phase, both original and reconstructed features are passed through a Dropout layer and then input to $C$, ensuring regularization and promoting better generalization. Finally, in the discrimination phase, both feature types are input into $\textit{D}$ to align feature distributions across domains by distinguishing between domain-specific and domain-invariant features.

In this setup, $g$ and $\textit{D}$ engage in a minimax game: $g$ extracts generalizable features from the data, while $\textit{D}$ distinguishes the features from the source and target domains. The discriminator loss can be computed as:
\begin{equation}
    \mathcal{L}_{D} = - \frac{1}{n}\sum_{x_{i}\in T } \mathcal{L}_{ce} (D(g(x_{i}^{*})), y_{i}^{d})
\end{equation}

where $n = n_{s} + n_{t}$, $y^{d}$ denotes the domain label, with $y^{d}=1$ indicating the source domain and $y^{d}=0$ indicating the target domain. The superscript * represents either $s$ (source domain) or $t$ (target domain).

To boost the adaptability of the model for the target domain, our framework emphasizes aligning the category-level feature space. This alignment needs to filter out task-independent nuisance factors while maintaining pure semantic information. 

A typical DAE is a self-training deep neural network architecture that does not leverage label information. Figure. 4 visualizes the process of reconstructing a corrupted input into a purified one. In our experiments, the input $x$ is corrupted by introducing Gaussian noise, with the extent of corruption being determined by a randomly generated ratio $v$.
This process removes information about the selected components from the input pattern, inducing the Auto-Encoder to subsequently learn to "fill" these introduced "gaps". 

The encoder operates on the corrupted data $\tilde{x}$ to encode it into the latent representation $y = f_{\theta}(\tilde{x}) = s(W\tilde{x} + b)$. Here, $\theta$ represents the parameters of the encoder. Subsequently, the decoder outputs the reconstructed data $z = g_{\theta'}(y) = s({W}' y + b)$ from the representation, where ${\theta}'$ denotes the parameters of the decoder.

The process of reconstruction is implemented through the following optimization:
\begin{equation}
    \mathop{\arg\min}\limits_{\theta,{\theta}' } \left [ L_{re}(X, g_{{\theta}'} (f_{\theta} (\tilde {X}))) \right ]
\end{equation}

Through learning the above reconstruction loss, the model can obtain advanced semantic information about the target domain dataset.

\section{Experiments}

\subsection{Setups}
To evaluate our proposed model GrabDAE, we conduct experiments on three public benchmarks.

\textbf{Office-Home}\cite{venkateswara2017deep}. Office-Home comprises 15,500 images spanning 65 classes and 4 distinct domains: Clipart (C), Art (A), Product (P), and Real-World (R). Each domain encompasses 65 categories.

\textbf{VisDA-2017}\cite{saenko2010adapting}. Designed to address the simulation-to-real-world transition, VisDA-2017 encompasses approximately 280K images distributed across 12 classes.

\textbf{Office31}\cite{peng2017visda}. Consisting of three subsets—Amazon (A), Webcam (W), and DSLR (D)—Office31 features images sourced from amazon.com for the Amazon subset, and images captured by a web camera and a DSLR camera for the Webcam and DSLR subsets, respectively. Overall, the dataset comprises 4,652 samples spread across 31 categories.

\begin{table*}[ht]
    \normalsize
    \renewcommand{\arraystretch}{1.3}
    \caption{Experiment Result in VisDA2017 Dataset}
    \label{Table performance_on_VisDA2017}
    \centering
    \resizebox{\textwidth}{!}{
    \begin{tabular}{c|ccccccccccccc}
    \hline
    Method                & plane & bicycle & bus  & car  & horse & knife & motor & person & plant & sktbrd & train & truck & Avg  \\ \hline
    ResNet(\textbf{CVPR 2015})\cite{targ2016resnet}   & 55.1  & 53.3    & 61.9 & 59.1 & 80.6  & 17.9  & 79.7  & 31.2   & 81.0    & 26.5   & 73.5  & 8.5   & 52.4 \\
    BNM(\textbf{CVPR 2020})\cite{cui2020towards}               & 89.6  & 61.5    & 76.9 & 55.0   & 89.3  & 69.1  & 81.3  & 65.5   & 90.0   & 47.3   & 89.1  & 30.1  & 70.4 \\
    
    Deit(\textbf{ICML 2021})\cite{touvron2021training}      & 98.2  & 73.0      & 82.5 & 62.0   & 97.3  & 63.5  & 96.5  & 29.8   & 68.7  & 86.7   & 96.7  & 23.6  & 73.2 \\
    
    CDTrans(\textbf{CVPR 2021})\cite{xu2021cdtrans}         & 97.1  & 90.5    & 82.4 & 77.5 & 96.6  & 96.1  & 93.6  & \textcolor{green}{88.6}   & \textcolor{blue}{97.9}  & 86.9   & 90.3  & \textcolor{green}{62.8}  & 88.4 \\

    AMRC(\textbf{TM2022}) \cite{jing2022adversarial}        & 94.1    & 86.5    & 77.9   & 58.7  & 92.5  & 91.8  & 82.3   & 80.6  & 89.5 & 87.4 & 84.9  & 58.7 & 82.1 \\
    TVT(\textbf{WACV 2023})\cite{yang2023tvt}                               & 82.9  & 85.6    & 77.5 & 60.5 & 93.6  & 98.2  & 89.4  & 76.4   & 93.6  & 92.0   & 91.7  & 55.7  & 83.1 \\
    PMTrans(\textbf{CVPR 2023})\cite{zhu2023patch}                             & \textcolor{green}{99.4}  & 88.3    & 88.1 & 78.9 & 98.8  & \textcolor{green}{98.3} & \textcolor{green}{95.8}  & 70.3   & 94.6  & \textcolor{green}{98.3}   & \textcolor{green}{96.3}  & 48.5  & 88.0 \\

    DAD(\textbf{TIP 2024}) \cite{peng2024unsupervised}            & 97.4  & 89.6    & \textcolor{blue}{92.2} & \textcolor{blue}{91.6} & 97.3  & 97.0 & 95.1  & \textcolor{blue}{89.8}   & \textcolor{green}{97.2}  & 96.9   & 93.7  & 42.5  & 90.0 \\
    
    DAMP(\textbf{CVPR 2024}) \cite{du2024domain}                            & 98.7  & \textcolor{green}{92.8}    & \textcolor{green}{91.7} & \textcolor{green}{80.1} & \textcolor{green}{98.9}  & 96.9 & 94.9  & 83.2   & 93.9  & 94.9   & 94.8  & \textcolor{blue}{\textbf{70.2}}  & \textcolor{green}{90.9} \\
    
    GrabDAE(\textbf{Ours})                    & \textcolor{blue}{\textbf{99.5$\uparrow$}}  & \textcolor{blue}{\textbf{96.2$\uparrow$}}    & \textbf{91.4} & \textbf{80.0}   & \textcolor{blue}{\textbf{99.4$\uparrow$}}  & \textcolor{blue}{\textbf{99.0$\uparrow$}}    & \textcolor{blue}{\textbf{96.9$\uparrow$}}  & \textbf{82.4}   & \textbf{96.1}  & \textcolor{blue}{\textbf{99.5$\uparrow$}}   & \textcolor{blue}{\textbf{96.7$\uparrow$}}  & \textbf{62.1}  & \textcolor{blue}{\textbf{91.6$\uparrow$}} \\ \hline
    \end{tabular}
    }
    \begin{tablenotes}
      \small
      \item Note: \textcolor{blue}{Blue} represents the highest value, while \textcolor{green}{green} signifies the second highest value.
    \end{tablenotes}
\end{table*}

\begin{table*}[ht]
    \normalsize
    \renewcommand{\arraystretch}{1.3}
    \caption{Experiment Result in OfficeHome Dataset}
    \label{Table performance_on_OfficeHome}
    \centering
    \resizebox{\textwidth}{!}{
    \begin{tabular}{c|ccccccccccccc}
    \hline
    Method                      & A-C                      & A-P                               & A-R                      & C-A                      & C-P                               & C-R                               & P-A                      & P-C                    & P-R                               & R-A                      & R-C                      & R-P                      & Avg                      \\ \hline
    ResNet(\textbf{CVPR 2015})\cite{targ2016resnet}      & 44.9                     & 66.3                              & 74.3                     & 51.8                     & 61.9                              & 63.9                              & 52.4                     & 39.1                   & 71.2                              & 63.8                     & 45.9                     & 77.2                     & 59.4                     \\
    FixBi(\textbf{CVPR 2021})\cite{na2021fixbi}                            & 58.1                     & 77.3                              & 80.4                     & 67.7                     & 79.5                              & 78.1                              & 65.8                     & 57.9                   & 81.7                              & 76.4                     & 62.9                     & 86.7                     & 72.7                     \\
    IFDMN(\textbf{TMM 2021})\cite{deng2021informative}                              & 61.2                     & 80.4                              & 82.7                     & 69.8                     & 76.5                              & 78                                & 69.2                     & 59.2                   & 84.1                              & 75.3                     & 61.9                     & 86.2                     & 73.7                     \\
    CDTrans(\textbf{CVPR 2021})\cite{xu2021cdtrans}           & 68.8                     & 85.0                              & 86.9                     & 81.5                     & 87.1                              & 87.3                              & 79.6                     & 63.3                   & 88.2                              & 82.0                       & 66.0                       & 90.6                     & 80.5                     \\
    
     Swin(\textbf{ICCV 2021})\cite{liu2021swin}       & 72.7                     & 87.1                              & 90.6                     & 84.3                     & 87.3                              & 89.3                              & 80.6                     & 68.6                   & 90.3                              & 84.8                     & 69.4                     & 91.3                     & 83.6                     \\
    SDAT(\textbf{CVPR 2022})\cite{rangwani2022closer}                                  & 70.8                     & 87.0                              & 90.5                     & 85.2                     & 87.3                              & 89.7                              & 84.1                     & 70.7                   & 90.6                              & 88.3                     & 75.5                     & 92.1                     & 84.3                     \\
    TVT(\textbf{WACV 2023})\cite{yang2023tvt}                                & 74.9                     & 86.8                              & 89.5                     & 82.8                     & 87.9                              & 88.3                              & 79.8                     & 71.9                   & 90.1                              & 85.5                     & 74.6                     & 90.6                     & 83.6                     \\
    MIC(\textbf{CVPR 2023})\cite{hoyer2023mic}                                  & 80.2                     & 87.3                              & 91.1                     & 87.2                     & 90.0                              & 90.1                              & 83.4                     & 75.6                   & 91.2                              & 88.6                     & 78.7                     & 91.4                     & 86.2                     \\
    
    PMTrans(\textbf{CVPR 2023})\cite{zhu2023patch}                         & \textcolor{green}{81.3}                     & 92.9                              & \textcolor{green}{92.8}                     & \textcolor{green}{88.4}                     & \textcolor{green}{93.4}                              & \textcolor{green}{93.2}                              & \textcolor{green}{87.9}                     & \textcolor{green}{80.4}                   & \textcolor{green}{93.0}                              & \textcolor{green}{89.0}                     & \textcolor{green}{80.9 }                    & \textcolor{green}{94.8 }                    & \textcolor{green}{89.0}                     \\
    
    ICON(\textbf{NeurIPS 2023})\cite{yue2023make}      & 63.3                     & 81.3                              & 84.5                     & 70.3                     & 82.1                              & 81.0                              & 70.3                     & 61.8                   & 83.7                              & 75.6                     & 68.6                     & 87.3                     & 75.8                     \\

    DAD(\textbf{TIP 2024})\cite{peng2024unsupervised}      & 62.5                     & 78.6                             & 83.0                     & 70.4                     & 79.2                              & 79.8                              & 70.2                     & 58.3                   & 83.1                              & 76.3                     & 63.5                     & 88.2                     & 74.4                     \\
    
    DAMP(\textbf{CVPR 2024}) \cite{du2024domain}     & 75.7                     & \textcolor{blue}{94.2}                             & 92.0                     & 86.3                     & \textcolor{blue}{94.2}                              & 91.9                              & 86.2                     & 76.3                   & 92.4                              & 86.1                     & 75.6                     & 94.0                     & 87.1                     \\
    GrabDAE(\textbf{Ours})              & \textcolor{blue}{\textbf{88.1$\uparrow$}}            & \textcolor{green}{\textbf{93.4}}                     & \textcolor{blue}{\textbf{95.3$\uparrow$}}            & \textcolor{blue}{\textbf{93.0$\uparrow$}}            & \textcolor{blue}{\textbf{94.2}}                     & \textcolor{blue}{\textbf{95.4$\uparrow$}}                     & \textcolor{blue}{\textbf{89.8$\uparrow$}}            & \textcolor{blue}{\textbf{86.9$\uparrow$}}          & \textcolor{blue}{\textbf{95.1$\uparrow$}}                     & \textcolor{blue}{\textbf{92.7$\uparrow$}}            & \textcolor{blue}{\textbf{89.0$\uparrow$}}            & \textcolor{blue}{\textbf{95.7$\uparrow$}}            & \textcolor{blue}{\textbf{92.4$\uparrow$}}            \\ \hline
    \end{tabular}
    }
    \begin{tablenotes}
      \small
      \item Note: \textcolor{blue}{Blue} represents the highest value, while \textcolor{green}{green} signifies the second highest value.
    \end{tablenotes}
\end{table*}

\begin{figure}[ht]
    \centering
    \begin{subfigure}{0.49\columnwidth}
        \centering
        \includegraphics[width=\linewidth]{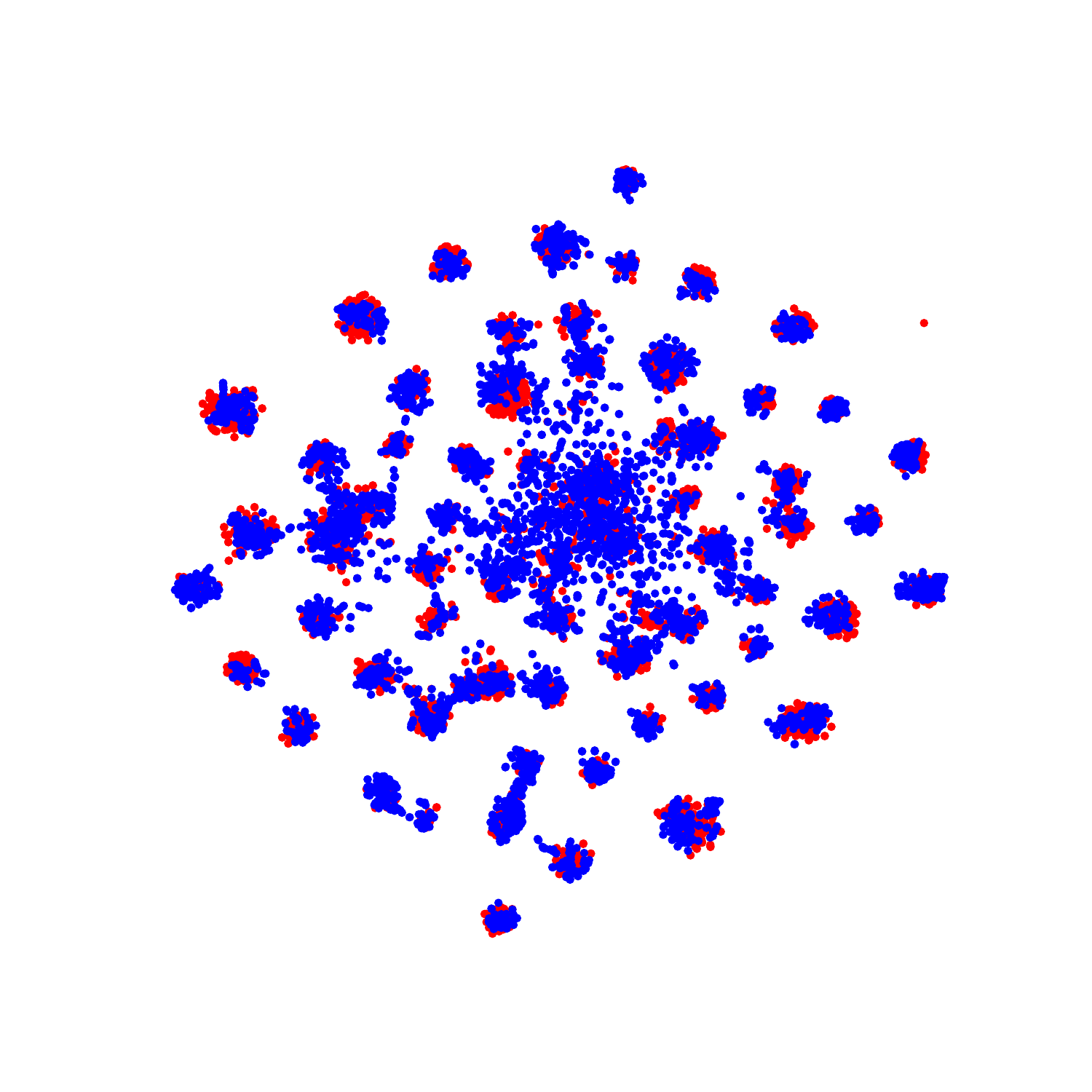}
        \label{SDAT}
        \caption{Global Alignment by SDAT}
    \end{subfigure}\hfill
    \begin{subfigure}{0.49\columnwidth}
        \centering
        \includegraphics[width=\linewidth]{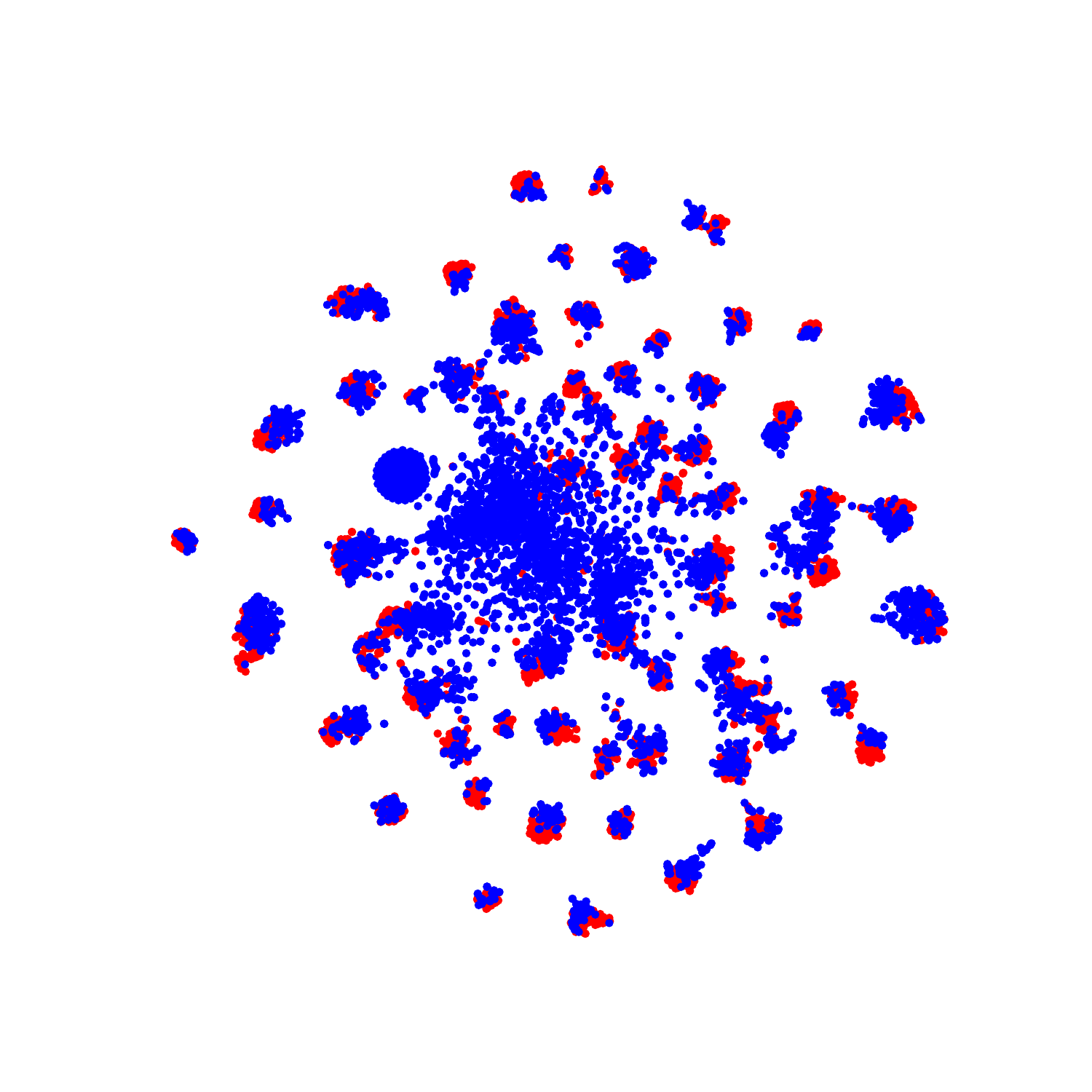}
        \label{Swin}
        \caption{Without Adaption}
    \end{subfigure}
    \begin{subfigure}{0.49\columnwidth}
        \centering
        \includegraphics[width=\linewidth]{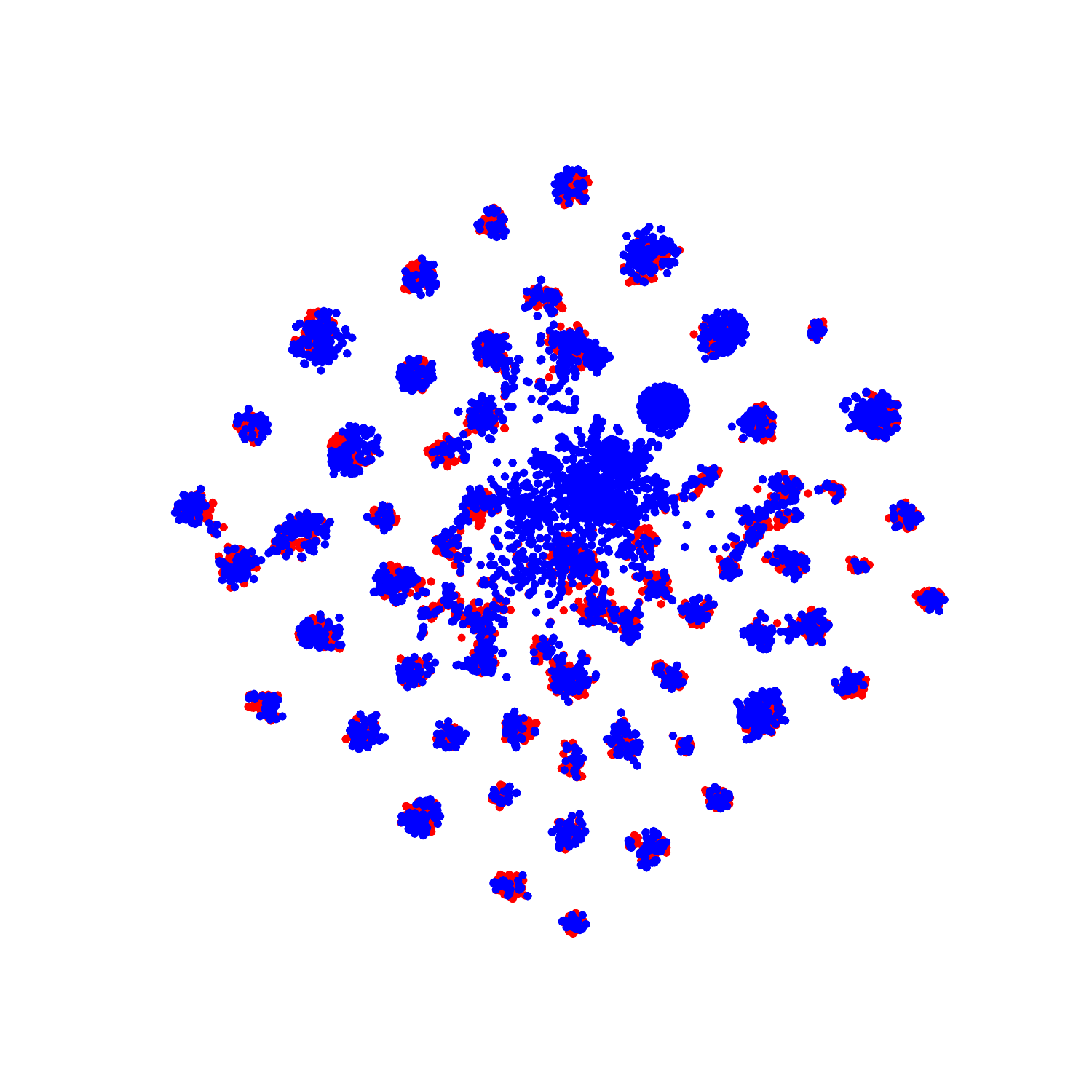}
        \label{MIC}
        \caption{Global Alignment by MIC}
    \end{subfigure}\hfill
    \begin{subfigure}{0.49\columnwidth}
        \centering
        \includegraphics[width=\linewidth]{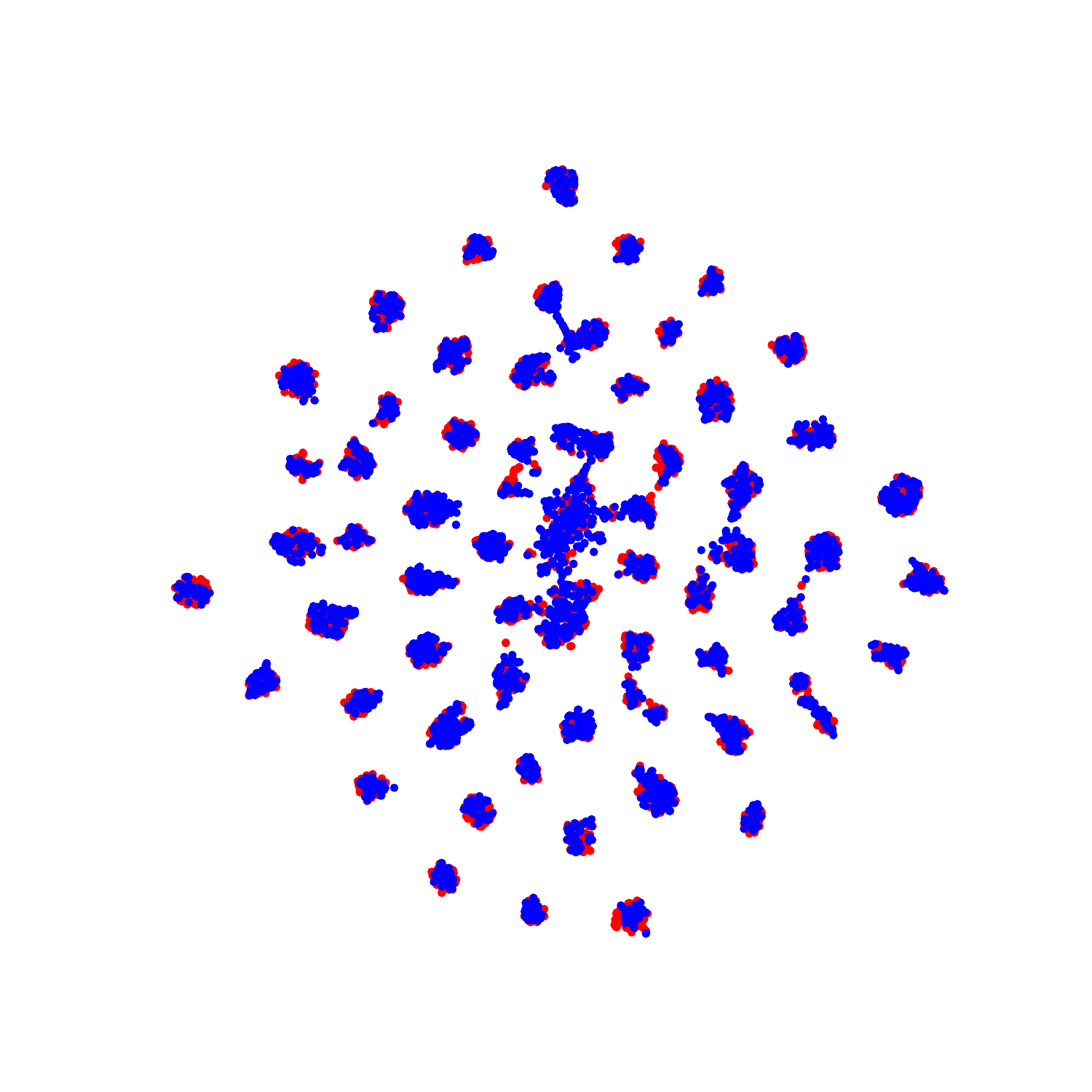}
        \label{GranDAE}
        \caption{Our Method}
    \end{subfigure}
    \caption{The t-SNE visualization of source(red) and target(blue) representation for Rw $\to$ Ar task from OfficeHome dataset. Learned features of our proposed method show a better separation of unknown samples from known classes than contemporary DA methods.}
    \label{fig}
\end{figure}

\begin{algorithm}[!h]
    \caption{Training Steps}
    \label{alg:AOS}
    
    \renewcommand{\algorithmicrequire}{\textbf{Input:}}
    \renewcommand{\algorithmicensure}{\textbf{Output:}}
    \renewcommand{\algorithmicwhile}{\textbf{While}} 
    \begin{algorithmic}[1]
        \REQUIRE Labeled source dataset:
        
        $S = {\{x_{i}^{s},y_{i}^{s}\mid x_{i}^{s}\in X_{s},y_{i}^{s}\in Y_{s}\}}$, unlabeled target dataset $D = {\{x_{i}^{t}\mid x_{i}^{t}\in X_{t}\}}$, teacher model $f_{t}$ and student model $f_{s}$, which include the feature extractor $g$, the DAE module, the discriminator $D$ and the Classifier $C$. 
        \ENSURE Classification result $\hat{Y}_{t}$    

        \WHILE{$t \leq MaxIter$}
            \STATE Obtain the masked images $X_{m}$ from target domain images $X_{t}$ utilizing the GrabMask Module.
            \STATE Predict the pseudo labels $P^{T}$ for target dataset $X_{t}$ by teacher model $f_{t}$ according to (11), while predict the labels $\hat y^{M}$ by student model $f_{s}$ according to (9).
            \STATE Compute target domain self-supervised loss $\mathcal{L}_{s}$ according to (10).
            \STATE Compute the cross-entropy loss $\mathcal{L}_{cls}$ according to (4) and domain adaption loss $\mathcal{L}_{DA}$ according to (6).
            \STATE Optimize the overall objective in (5) through stochastic gradient descent.
        \ENDWHILE
        
        \RETURN Outputs
    \end{algorithmic}
    
\end{algorithm}

\subsection{Experiment Protocols}
As shown in Fig. 3, the architecture of GrabDAE consists of four components: the domain discriminator $D$, the feature extractor $g$, the DAE module, and the classifier $C$. The feature extractor $g$ is implemented using the Swin Transformer. For classification, a conventional softmax classifier is employed, while the domain discriminator is implemented through Conditional Domain Adversarial Networks (CDAN)\cite{long2018conditional}. The DAE module is designed with three primary sub-components: a noise injection stage, followed by dedicated encoding and decoding processes.

The implementation of GrabDAE is facilitated by Pytorch, with optimization achieved via the Stochastic Gradient Descent (SGD) algorithm with a momentum of 0.9 and weight decay of 1e-4. In all experiments, we utilize the Swin-L as the backbone for our GrabDAE. The initial learning rate is set to 1e-3 with a batch size of 32, and we train the GrabDAE model for 30 epochs. Notably, for the VisDA-2017 dataset, this learning rate is adjusted to 5e-4.

\subsection{Results}
In the task of image classification in UDA, the model is trained with the source data and subsequently evaluated using the target data. 

\textbf{Results on VisDA-2017.} As shown in Table \ref{Table performance_on_VisDA2017}, our method reaches 91.6\% accuracy and outperforms the \textbf{baseline(CVPR 2024)} by 0.7\%. Notably, for some challenging categories, such as "bicycle," our method consistently realizes a significant performance improvement, increasing from 92.8\% to 96.2\%, an enhancement of 3.4\%.

\textbf{Results on OfficeHome.} As shown in Table \ref{Table performance_on_OfficeHome}, our GrabDAE framework achieves noticeable performance gains and surpasses TVT, DAMP, and MIC by a large margin. Importantly, our GrabDAE achieves an improvement of more than 3.4\% accuracy over the SOTA method PMTrans, yielding 92.4\% accuracy. Incredibly, GrabDAE almost surpasses the SOTA methods in all subtasks, which demonstrates the strong ability of GrabDAE to alleviate the large domain gap.

\textbf{Results on Office-31.} Table \ref{Table performance_on_Office31} shows the quantitative results of various models in different tasks. Overall, our GrabDAE achieves superior performance in most tasks, reaching 95.6\% accuracy and surpassing SOTA methods. Specifically, GrabDAE demonstrates significant improvements over other methods, with an accuracy increase of +0.3\% over PMTrans, +1.7\% over TVT, and +3.0\% over CDTrans.

\begin{table*}[ht]
    \normalsize
    \renewcommand{\arraystretch}{1.3}
    \caption{Experiment Result in Office31 Dataset}
    \label{Table performance_on_Office31}
    \centering
    \begin{tabular}{c|ccccccc}
    \hline
    Method           & A→W           & D→W           & W→D          & A→D           & D→A           & W→A           & Avg           \\ \hline
    ResNet(\textbf{CVPR 2015})\cite{targ2016resnet}       & 68.9          & 68.4          & 62.5         & 96.7          & 60.7          & 99.3          & 76.1          \\
    CWDAN(\textbf{TMM 2019})\cite{yan2019weighted}       & 73.5          & 97.8          & 99.4         & 67.8          & 54.8          & 54.5          & 74.6          \\
    BNM(\textbf{CVPR 2020})\cite{cui2020towards}                                & 91.5          & 98.5          & \textcolor{blue}{\textbf{100}}          & 90.3          & 70.9          & 71.6          & 87.1          \\
    
    IFDMN(\textbf{TMM 2021})\cite{deng2021informative}                       & 95.4          & 99.3          & \textcolor{blue}{\textbf{100}}          & 95.9          & 79.3          & 77.8          & 91.3          \\
    FixBi(\textbf{CVPR 2021})\cite{na2021fixbi}                & 96.1          & 99.3          & \textcolor{blue}{\textbf{100}}          & 95.0          & 78.7          & 79.4          & 91.4          \\ 
    
    CDTrans(\textbf{CVPR 2021})\cite{xu2021cdtrans}          & 96.7          & 99.0            & \textcolor{blue}{\textbf{100}}          & 97.0            & 81.1          & 81.9          & 92.6          \\
    ViT(\textbf{ICLR 2021})\cite{dosovitskiy2020image}                          & 91.2          & 99.2          & \textcolor{blue}{\textbf{100}}          & 90.4          & 81.1          & 80.6          & 91.1          \\
    Swin(\textbf{ICCV 2021})\cite{liu2021swin}        & 97.0          & 99.2          & \textcolor{blue}{\textbf{100}}          & 95.8          & 82.4          & 81.8          & 92.7          \\
    CDCL(\textbf{TMM 2022})\cite{wang2022cross}       & 96.0          & 99.2          & \textcolor{blue}{\textbf{100}}         & 96.0          & 76.4          & 74.1          & 90.6          \\
    SSRT(\textbf{CVPR 2022})\cite{iftekhar2022look}              & 97.7          & 99.2          & \textcolor{blue}{\textbf{100}}          & 98.6          & 83.5          & 82.2          & 93.5          \\

    TVT(\textbf{WACV 2023})\cite{yang2023tvt}             & 96.4          & \textcolor{green}{\textbf{99.4}}          & \textcolor{blue}{\textbf{100}}          & 96.4          & 84.9          & 86.0          & 93.9          \\
    PMTrans(\textbf{CVPR 2023})\cite{zhu2023patch}                & \textcolor{blue}{99.5}          & \textcolor{green}{\textbf{99.4}}          & \textcolor{blue}{\textbf{100}}          & \textcolor{blue}{\textbf{99.8} }         & \textcolor{green}{86.7}          & \textcolor{green}{86.5}          & \textcolor{green}{95.3}          \\ 

    DAD(\textbf{TIP 2024})\cite{peng2024unsupervised}             & \textcolor{green}{98.5}         & \textcolor{blue}{\textbf{99.5}}          & \textcolor{blue}{\textbf{100}}          & 95.6          & 81.4         & 82.2          & 92.8          \\
    
    GrabDAE(\textbf{Ours})        & \textcolor{blue}{\textbf{99.5}} & 99.2 & \textcolor{blue}{\textbf{100}} & \textcolor{green}{99.2} & \textcolor{blue}{\textbf{87.5$\uparrow$}}  & \textcolor{blue}{\textbf{87.9$\uparrow$}} & \textcolor{blue}{\textbf{95.6$\uparrow$}} \\ \hline
    \end{tabular}
    \begin{tablenotes}
      \small
      \item Note: \textcolor{blue}{Blue} represents the highest value, while \textcolor{green}{green} signifies the second highest value. Our model did not attain the state-of-the-art (SOTA) in every task on the Office31 dataset, possibly due to the limited potential for improvement resulting from the close resemblance of data between the two domains within the dataset. Nonetheless, in the more challenging tasks, our model consistently achieved SOTA, showcasing its exceptional generalization capabilities.
    \end{tablenotes}
\end{table*}

\begin{table*}[ht]
    \normalsize
    \renewcommand{\arraystretch}{1.3}
    \caption{Mask Ablation Experiment Result in OfficeHome Dataset}
    \label{Ablation study about mask approach}
    \centering
    \begin{tabular}{c|ccccccccccccc}
    \hline
    Approach     & A-C           & A-P           & A-R           & C-A           & C-P           & C-R           & P-A           & P-C           & P-R           & R-A           & R-C           & R-P           & Avg           \\ \hline
    None     & 83.6          & 90.9          & 93.8          & 90.4          & 92.7          & 93.6          & 88.0          & 82.8          & 93.9          & 90.7          & 85.5          & 94.7          & 90.1          \\
    MaskRNN\cite{hu2017maskrnn}  & 72.2          & 84.7          & 92.6          & 73.6          & 71.7          & 91.0          & 67.5          & 53.4          & 92.7          & 67.3          & 67.5          & 94.0          & 77.4         \\
    Saliency\cite{hou2007saliency} & 83.9          & 92.0          & 93.1          & 90.7          & 92.7          & 93.4          & 87.9          & 83.4          & 93.7          & 90.7          & 85.6          & 94.6          & 90.1          \\
    GrabMask & \textcolor{blue}{\textbf{88.1$\uparrow$}} & \textcolor{blue}{\textbf{93.4$\uparrow$}} & \textcolor{blue}{\textbf{95.3$\uparrow$}} & \textcolor{blue}{\textbf{93.0$\uparrow$}} & \textcolor{blue}{\textbf{94.$\uparrow$2}} & \textcolor{blue}{\textbf{95.4$\uparrow$}} & \textcolor{blue}{\textbf{89.8$\uparrow$}} & \textcolor{blue}{\textbf{86.9$\uparrow$}} & \textcolor{blue}{\textbf{95.1$\uparrow$}} & \textcolor{blue}{\textbf{92.7$\uparrow$}} & \textcolor{blue}{\textbf{89.0$\uparrow$}} & \textcolor{blue}{\textbf{95.7$\uparrow$}} & \textcolor{blue}{\textbf{92.4$\uparrow$}} \\ \hline
    \end{tabular}
    \begin{tablenotes}
      \small
      \item Note: \textcolor{blue}{Blue} represents the highest value.
    \end{tablenotes}
\end{table*}

\begin{table*}[ht]
    \normalsize
    \renewcommand{\arraystretch}{1.3}
    \caption{Ablation Study: Accuracy(\%) of the proposed method and its different variants in OfficeHome Dataset}
    \label{tab_fwsc}
    \centering
    \begin{tabular}{ccc|ccccccccccccc}
    \hline
    \multicolumn{1}{l}{$L_{cls}$} & \multicolumn{1}{l}{$L_{T}$} & \multicolumn{1}{l|}{$L_{re}$} & A-C           & A-P           & A-R           & C-A         & C-P           & C-R           & P-A           & P-C           & P-R           & R-A           & R-C         & R-P           & Avg           \\ \hline
    $\surd$                         &                        &                          & 78.7          & 89.6          & 91.0            & 87.4        & 88.9          & 90.9          & 87.3          & 81.9 & 92.3          & 89.0            & 85.9        & 93.3          & 88.8          \\
    $\surd$                          & $\surd$                       &                          & 85.7          & 92.8          & 93.9          & 91.3        & 93.2          & 94.0            & 88.8          & 85.3          & 94.0            & 91.6          & 87.3        & 95.0            & 91.1          \\
    $\surd$                          &                        & $\surd$                          & 83.6          & 90.9          & 93.8          & 90.4        & 92.7          & 93.6          & 88.0            & 82.8          & 93.9          & 90.7          & 85.5        & 94.7          & 90.1          \\
    $\surd$                          & $\surd$                       & $\surd$                          & \textcolor{blue}{\textbf{88.1$\uparrow$ }} & \textcolor{blue}{\textbf{93.4$\uparrow$}} & \textcolor{blue}{\textbf{95.3$\uparrow$}} & \textcolor{blue}{\textbf{93$\uparrow$}} & \textcolor{blue}{\textbf{94.2$\uparrow$}} & \textcolor{blue}{\textbf{95.4$\uparrow$}} & \textcolor{blue}{\textbf{89.8$\uparrow$}} & \textcolor{blue}{\textbf{86.9$\uparrow$}}     & \textcolor{blue}{\textbf{95.1$\uparrow$}} & \textcolor{blue}{\textbf{92.7$\uparrow$}} & \textcolor{blue}{\textbf{89.0$\uparrow$}} & \textcolor{blue}{\textbf{95.7$\uparrow$}} & \textcolor{blue}{\textbf{92.4$\uparrow$}} \\ \hline
    
    \end{tabular}
    \begin{tablenotes}
      \small
      \item Note: \textcolor{blue}{Blue} represents the highest value.
    \end{tablenotes}
\end{table*}

\subsection{Ablation Study}

To unveil the individual contributions of the GrabDAE framework's components, we conducted comprehensive ablation studies. These experiments were designed to validate the effectiveness of our approach in Unsupervised Domain Adaptation for image classification. Using the Swin Transformer with SDAT \cite{rangwani2022closer} as our baseline, we explored the impact of the innovative Grab-Mask module and other key elements of GrabDAE.

\subsubsection{Evaluating the Grab-Mask Module}

A focal point of our study was the Grab-Mask module. We compared it against alternatives like MaskRNN \cite{hu2017maskrnn} and saliency detection \cite{montabone2010human} to highlight the comparative advantages of the Grab-Mask module in domain adaptation. Table \ref{Ablation study about mask approach} shows that incorporating Grab-Mask into GrabDAE significantly enhances accuracy by 2.3\% over the baseline and by 15.1\% over the variant using MaskRNN. This improvement showcases Grab-Mask's ability to effectively focus the model on relevant visual information, reducing the impact of background distractions.

The superior performance of Grab-Mask can be attributed to its saliency-aware preprocessing mechanism, which effectively enhances the model's focus on relevant visual features and reduces the influence of irrelevant background elements. Unlike MaskRNN, which may introduce noise or inaccuracies during the masking process, Grab-Mask leverages Gaussian Mixture Models to perform soft segmentation, providing a more reliable and accurate representation of salient regions within images. Additionally, compared to a standalone saliency module, Grab-Mask integrates seamlessly into the GrabDAE framework, enabling more effective feature extraction and domain adaptation.

\subsubsection{Component-Wise Effectiveness}

We further dissected GrabDAE's performance by examining different combinations of classification loss ($L_{cls}$), self-supervised loss ($L_{s}$), and reconstruction loss ($L_{re}$). The collective application of these components, as seen in Table \ref{tab_fwsc}, yielded the highest average accuracy of 92.4\%. This configuration outperformed others, emphasizing the crucial role of reconstruction loss in enhancing feature quality and minimizing domain differences.

\subsubsection{t-SNE Visualization of Model Performance}

To elucidate the efficacy of our training scheme on domain-specific feature distribution, we utilized t-SNE to visualize the embeddings of the target domain within the OfficeHome dataset, particularly focusing on the \textit{Rw} $\to$ \textit{Ar} UDA task. Figure. 5 offers a comparative insight into the feature representations learned through our GrabDAE framework versus those from other alternative models. Figure. 5(b) shows the visualization of features without adaptation, while Figure. 5(a) and (c) depict features after global alignment from other models-(c) illustrate the feature embeddings from both the source and target domains as learned by the competing models. Figure. 5(d) showcases the embeddings derived from GrabDAE and reveals a more cohesive alignment of domain distributions, which demonstrates the GrabDAE's capability in mitigating the divergence between domains and aligning category-level features more effectively. This visual comparison underscores the superior adaptability and performance of our method in handling domain shifts, highlighting its potential in fostering robust domain adaptation.

In summary, this ablation study clearly outlines the essential roles played by the Grab-Mask module and GrabDAE's loss components in achieving superior domain adaptation performance. The empirical evidence supports our framework's efficacy in addressing domain shifts, paving the way for future research to expand our methodology to a broader range of tasks and challenges in domain adaptation.

\section{Conclusion}

In this study, we presented GrabDAE, a novel Unsupervised Domain Adaptation (UDA) framework for image classification. This framework addresses domain transfer challenges during training by purifying features through the reconstruction of masked features using the Denoise Auto-Encoder (DAE) module. Our empirical evaluations demonstrate that GrabDAE achieves superior category-level alignment and competitive, often superior, adaptation performance compared to more complex methods. Moreover, the applications of our approach extend beyond object classification to areas such as 3D visual learning, object detection, and semantic segmentation, which are also affected by domain shifts.

Future work could explore integrating additional self-supervised learning techniques to enhance feature extraction and domain alignment. Another promising direction is applying the extension of the GrabDAE framework to different Unsupervised Domain Adaptation (UDA) settings, such as open-set domain adaptation\cite{chang2023mind,li2023adjustment}, where the target domain contains unknown classes not present in the source domain. Extending the framework to handle more complex and heterogeneous datasets, including multi-modal data, may also broaden its applicability and robustness. Extending GrabDAE to handle more complex and heterogeneous datasets, including multi-modal data, would also broaden its applicability and robustness. Overall, GrabDAE offers a strong foundation for UDA in image classification, with substantial potential for adaptation to diverse domain-shifted tasks.

\textcolor{blue}{\textbf{We would like to release our source code.}}

\section*{Acknowledgments}
This project is jointly supported by National Natural Science Foundation of China (Nos. 61003143, 52172350),Guangdong Basic and Applied Research Foundation (Nos.2021B1515120032, 2022B1515120072), Guangzhou Science and Technology Plan Project (No.2024B01W0079), Nansha Key RD Program (No.2022ZD014), Science and Technology Planning Project of Guangdong Province (No.2023B1212060029).

\bibliographystyle{IEEEtran}
\bibliography{refer}

\vfill

\end{document}